\documentclass[lettersize,journal]{IEEEtran}
\usepackage{amsmath,amsfonts}
\usepackage{algorithmic}
\usepackage{algorithm}
\usepackage{array}
\usepackage[caption=false,font=normalsize,labelfont=sf,textfont=sf]{subfig}
\usepackage{textcomp}
\usepackage{stfloats}
\usepackage{url}
\usepackage{verbatim}
\usepackage{graphicx}
\usepackage{booktabs}
\usepackage{cite}
\usepackage{multirow}
\usepackage[colorlinks=true, linkcolor=blue, citecolor=blue, urlcolor=blue]{hyperref}
\begin{document}

\title{Exploring Remote Physiological Signal Measurement under Dynamic Lighting Conditions at Night: Dataset, Experiment, and Analysis}

\author{
    Zhipeng Li, 
    Kegang Wang,
    Hanguang Xiao, 
    Xingyue Liu,
    Feizhong Zhou,
    Jiaxin Jiang,
    Tianqi Liu
    
    

}
\date{}


\maketitle

\begin{abstract}
Remote photoplethysmography (rPPG) is a non-contact technique for measuring human physiological signals. Due to its convenience and non-invasiveness, it has demonstrated broad application potential in areas such as health monitoring and emotion recognition. In recent years, the release of numerous public datasets has significantly advanced the performance of rPPG algorithms under ideal lighting conditions. However, the effectiveness of current rPPG methods in realistic nighttime scenarios with dynamic lighting variations remains largely unknown. Moreover, there is a severe lack of datasets specifically designed for such challenging environments, which has substantially hindered progress in this area of research. To address this gap, we present and release a large-scale rPPG dataset collected under dynamic lighting conditions at night, named DLCN. The dataset comprises approximately 13 hours of video data and corresponding synchronized physiological signals from 98 participants, covering four representative nighttime lighting scenarios. DLCN offers high diversity and realism, making it a valuable resource for evaluating algorithm robustness in complex conditions. Built upon the proposed Happy-rPPG Toolkit, we conduct extensive experiments and provide a comprehensive analysis of the challenges faced by state-of-the-art rPPG methods when applied to DLCN. The dataset and code are publicly available at: \url{https://github.com/dalaoplan/Happp-rPPG-Toolkit}.
\end{abstract}


\section{Introduction}
\IEEEPARstart{P}hysiological signals such as heart rate serve as critical indicators of an individual’s health status. Currently, the measurement of such signals primarily relies on electrocardiography (ECG) and photoplethysmography (PPG). However, both techniques require direct contact between sensors and the human skin, which can present considerable limitations in specific scenarios, such as sleep monitoring, neonatal care, and elderly assistance, where contact-based measurements may cause inconvenience or even discomfort~\cite{1}. Remote photoplethysmography (rPPG) offers a non-contact and cost-effective alternative. This technique enables the remote measurement of stable physiological signals by capturing facial videos using a standard RGB camera. In recent years, due to its practicality and ease of use, rPPG has been widely applied across various domains, including health monitoring~\cite{2}, emotion recognition~\cite{3}, anti-spoofing detection~\cite{4}, and driver safety~\cite{5}.

\begin{figure}[t]
	\centering 
	\includegraphics[width=0.48\textwidth]{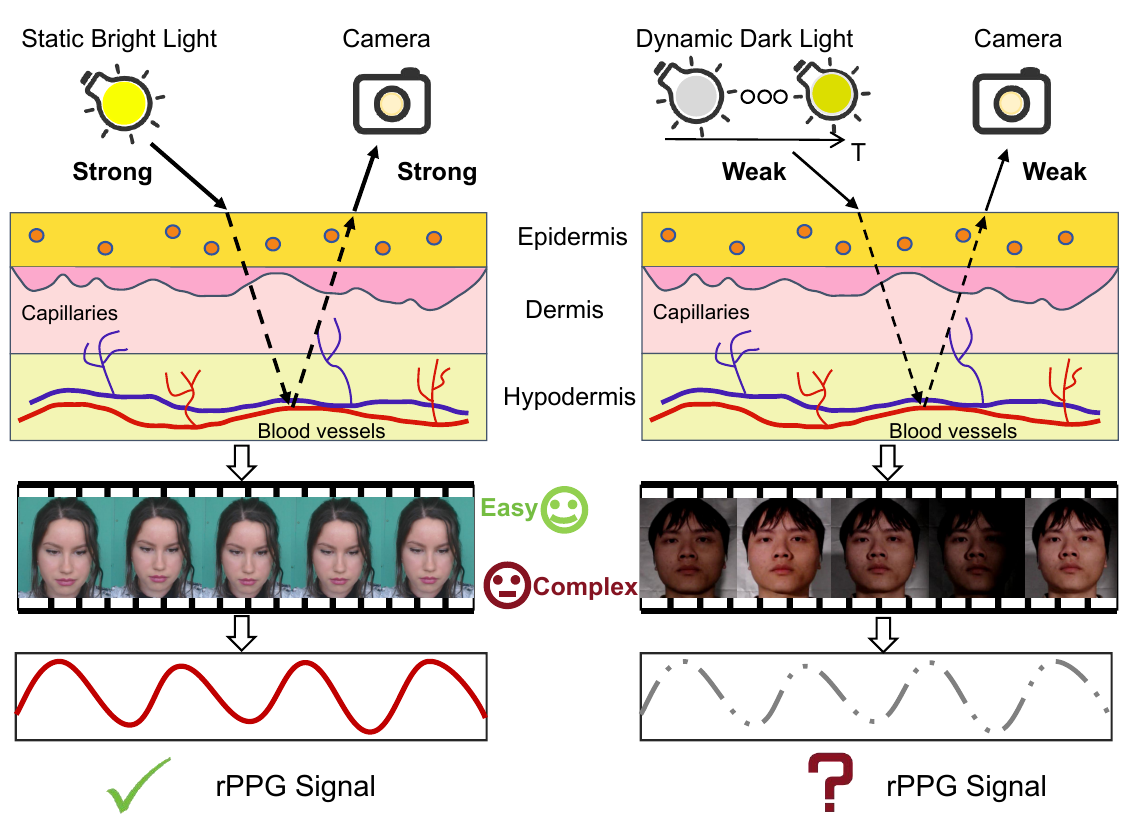}
	\caption{Comparison between current mainstream datasets and rPPG signal acquisition under night-time dynamic lighting conditions. Under sufficient and stable illumination, rPPG signals can be extracted reliably from facial videos. However, in night-time dynamic lighting environments, the extraction process remains highly uncertain and challenging.} 
	\label{1rPPGmethod} 
\end{figure}

\par{With the continuous advancement of research, the field of remote photoplethysmography (rPPG) has witnessed significant progress. Existing rPPG methods can be broadly categorized into two groups: traditional handcrafted approaches and data-driven deep learning methods~\cite{6}. Traditional methods, such as CHROM~\cite{7}, POS~\cite{8}, and ICA~\cite{9}, primarily rely on predefined rules, including color space transformations, illumination compensation, and signal filtering. The performance of these methods largely depends on human expertise and parameter tuning. However, in practical applications, facial videos are often affected by various sources of interference, such as head movements, facial expressions, and fluctuations in ambient lighting~\cite {10}. These factors greatly compromise the stability and accuracy of traditional methods. In particular, under complex lighting conditions, extracting high-quality physiological signals remains a major challenge. In contrast, deep learning–based methods leverage large-scale data to automatically learn spatiotemporal features, demonstrating stronger robustness and generalization capability, thereby enabling better handling of complex interferences. For example, models such as EfficientPhys~\cite{11}, PhysNet~\cite{12}, and PhysFormer~\cite{13} have introduced 2D convolutional neural networks (2D CNNs), 3D convolutional neural networks (3D CNNs), and attention mechanisms, achieving state-of-the-art performance across multiple public datasets.}

\par{However, it is important to note that most existing public rPPG datasets are collected under relatively ideal conditions, featuring uniform and stable lighting. Research on rPPG signal extraction in dynamic lighting environments, especially in complex nighttime scenarios, remains limited. Previous efforts, such as BUAA-MIHR~\cite{14}, introduced varying levels of low illumination to examine the feasibility of rPPG under nighttime conditions. However, this work focused solely on static variations in light intensity. Subsequent datasets such as VIPL~\cite{15} and MMPD~\cite{10} incorporated head movements to simulate lighting fluctuations caused by relative motion. Nevertheless, since these datasets were still collected in controlled laboratory settings, the resulting illumination changes were limited in scale and failed to accurately reflect the severe and complex lighting disturbances commonly observed in real-world nighttime environments.}

\par{In practice, lighting variations in real nighttime scenarios are far more complex and unpredictable. Factors such as flashing advertisements, vehicle headlights, and occlusions from surrounding buildings can introduce significant interference in rPPG signal extraction~\cite{5}. These dynamic lighting disturbances are not sufficiently represented in existing datasets, making it difficult to reliably evaluate model performance under such challenging conditions (as illustrated in Fig.~\ref{1rPPGmethod}). Consequently, the lack of complexity, particularly in terms of nighttime dynamic lighting conditions, has become a major bottleneck restricting the further development and real-world deployment of rPPG technology.}

\par{
To address the lack of nighttime dynamic lighting conditions in existing rPPG datasets, we introduce a novel and publicly available dataset—DLCN (Dynamic Lighting Conditions at Night). This dataset comprises 784 video samples collected from 98 participants across four representative nighttime lighting scenarios. To enhance physiological diversity, recordings were conducted under both resting conditions and post-exercise states, resulting in a broader range of heart rates. This diversity is expected to improve the generalizability and practical value of rPPG algorithms.
}

\par{The main contributions of this work are as follows:}

\par{(1) We construct and release the first rPPG dataset specifically focused on nighttime dynamic lighting conditions. To the best of our knowledge, DLCN is the first publicly available dataset that systematically targets complex, real-world nighttime lighting variations, addressing a critical gap in current rPPG research resources.
}

\par{(2) We conduct a comprehensive evaluation of both traditional and deep learning–based rPPG methods under nighttime dynamic lighting conditions, thoroughly analyzing their adaptability and limitations in such challenging scenarios. The results provide valuable benchmarks and insights for the design, improvement, and optimization of future rPPG algorithms.}

\par{(3) We develop and release a lightweight algorithmic toolkit for rPPG—Happy-rPPG Toolkit. This toolkit integrates multiple mainstream methods and supports training and evaluation on both DLCN and other public datasets, facilitating algorithm reproducibility and comparative studies within the community.}

\par{The remainder of this paper is organized as follows: Section~\ref{S2RelatedWork} reviews related work on remote physiological signal measurement, publicly available datasets, and rPPG methods under varying lighting conditions. Section~\ref{S3Method} provides a detailed description of the construction of the proposed DLCN dataset and the associated challenges. Section~\ref{S4Experiments} presents comprehensive experimental evaluations, followed by an in-depth analysis and discussion of the results in Section~\ref{S5Discussion}. Finally, Section~\ref{S6Conclusion} concludes the paper.}

\section{Related Work}
\label{S2RelatedWork}
\subsection{rPPG Measurement}
\subsubsection{Traditional Model-based rPPG Methods}
\par{Early rPPG methods primarily relied on traditional handcrafted modeling strategies, with research efforts focused on extracting pulse-related information from RGB color channels. One of the earliest works, GREEN by Verkruysse et al.~\cite{16}, analyzed signals from the three color channels of video frames and found that the green channel contained the most prominent pulse-related components. Building on this insight, subsequent studies introduced blind source separation techniques to enhance signal extraction. Methods such as Independent Component Analysis (ICA)~\cite{9} and Principal Component Analysis (PCA)~\cite{17} assume statistical independence among source signals and attempt to isolate the periodic pulse component from mixed signals. However, in real-world applications, these methods tend to suffer from significant performance degradation when motion artifacts or illumination variations exhibit periodic characteristics themselves, leading to reduced robustness. To address this limitation, later research proposed color space–based pulse extraction methods. These methods aim to project raw skin color signals into an optimized subspace that enhances the pulse component while suppressing noise. For example, PBV~\cite{18} projects color variations onto a predefined pulse direction vector to extract the relevant signal component directly. CHROM~\cite{7} removes specular reflection components unrelated to the pulse and projects chrominance information onto a plane orthogonal to the direction of specular variation, thereby strengthening the pulse signal. 2SR~\cite{19} constructs a temporally rotating skin-color subspace and leverages the periodicity of its rotational behavior for pulse extraction. In addition, invariant features in local color variations have been introduced to improve robustness under complex disturbances. For instance, LGI~\cite{20} utilizes invariances within local group transformations, reordering the blood volume pulse signals in a vector space. By exploiting the temporal consistency of heart rate within local time windows, this method enhances pulse signal detectability and resistance to noise.}

\subsubsection{Deep Learning-based rPPG Methods}
\par{
With the rapid development of deep learning, an increasing number of deep neural network–based rPPG methods have been proposed, significantly improving the accuracy and robustness of pulse signal extraction. Among them, end-to-end deep learning frameworks constitute a prominent class of approaches. DeepPhys, proposed by Chen et al.~\cite{21}, adopts a dual-branch 2D convolutional neural network (CNN) that takes raw frames and frame-difference images as input, effectively integrating spatial and short-term dynamic information for pulse extraction. To address the limited temporal modeling capability of 2D CNNs, TS-CAN~\cite{22} incorporates the Temporal Shift Module (TSM)~\cite{23}, which enhances the network's ability to capture long-range dependencies in temporal sequences, thereby improving the stability of signal estimation. In terms of architectural optimization, BigSmall~\cite{24}, inspired by the SlowFast network~\cite{25}, introduces fast and slow branches to handle information at different temporal frequencies, significantly reducing computational cost without sacrificing accuracy. EfficientPhys~\cite{11} further simplifies the dual-branch design into a single-branch structure, achieving improved computational efficiency while maintaining competitive performance. PhysNet, proposed by Yu et al.~\cite{12}, employs a 3D CNN architecture that exhibits strong spatiotemporal modeling capabilities across video sequences, yielding superior performance. However, the high computational cost of 3D CNNs has prompted researchers to explore lightweight alternatives that preserve modeling capacity. Methods such as RTrPPG~\cite{26}, LSTC-rPPG~\cite{27}, JAMSNet~\cite{28}, and iBVPNet~\cite{29} reduce model complexity through techniques like multi-scale fusion, module reconfiguration, and network pruning. Recently, Transformer-based architectures have also been introduced to the rPPG domain. PhysFormer~\cite{13} integrates the Vision Transformer into the rPPG pipeline and demonstrates stable performance across multiple benchmark datasets. Further advancements include RhythmFormer~\cite{30}, which employs a sparse attention mechanism, and Spiking-PhysFormer~\cite{31}, which introduces spiking neural networks (SNNs) to reduce computational cost while preserving accuracy. In addition, another line of research focuses on non–end-to-end methods that extract physiological features by constructing spatiotemporal signal maps (STMaps) of facial regions. Representative examples include SynRhythm~\cite{32}, RhythmNet~\cite{33}, and Dual-GAN~\cite{34}. These approaches exhibit strong robustness against illumination and facial expression variations but rely on sophisticated preprocessing steps to generate high-quality STMaps.
}

\begin{table*}[t]
\centering
\caption{Comparison of dataset.}
\label{table1}
\begin{tabular}{lcccccc}
\toprule
\textbf{Dataset} & \textbf{Frames} & \textbf{Subjects} & \textbf{Resolution / FPS} & \textbf{Label} & \textbf{Lighting Conditions} & \textbf{HR Range} \\
\midrule
UBFC-rPPG  & 57,420    & 42  & 640$\times$480 / 30 & PPG, HR      & Bright (Static)                  & 60–140   \\
PURE       & 168,120   & 10  & 640$\times$480 / 30 & PPG, SpO$_2$ & Dim (Static)                     & 42–148   \\
COHFACE    & 192,000   & 40  & 640$\times$480 / 20 & PPG, HR      & Bright \& Dim (Static)           & 45–97    \\
DLCN       & 1,411,200 & 98  & 640$\times$480 / 30 & PPG, HR, SpO$_2$ & Bright \& Dim (Static + Dynamic) & 45–154   \\
\bottomrule
\end{tabular}
\end{table*}

\subsection{Public rPPG Datasetss}
\par{Several high-quality public datasets have been released to support the development of rPPG algorithms, including PURE~\cite{2}, UBFC-rPPG~\cite{35}, COHFACE~\cite{36}, MMPD~\cite{10}, VIPL-HR~\cite{15}, BUAA-MIHR~\cite{14}, CHILL~\cite{37}, RLAP~\cite{38}, and MR-NIRP~\cite{5}. For representativeness and comparability, this study selects the three most widely used datasets for detailed analysis and discussion. A summary of their specifications is presented in Table~\ref{table1}.}

\par{PURE~\cite{2} consists of 60 one-minute videos collected from 10 subjects performing six different tasks (e.g., resting, speaking, and moving), along with corresponding physiological signals. Videos were recorded using an eco274CVGE camera at a resolution of $640\times480$ and a frame rate of 30 fps. Ground-truth signals were captured using a CMS50E pulse oximeter clipped to the finger. The recordings were conducted under indoor natural lighting with generally dim illumination. The only light variation originates from minor natural light fluctuations through a window, and illumination can be considered nearly stable within each one-minute clip.}

\par{UBFC-rPPG~\cite{35} includes video recordings of 42 subjects, accompanied by corresponding PPG signals and heart rate (HR) labels. Videos were recorded using a Logitech C920 HD Pro webcam at $640\times480$ resolution and 30 fps. PPG signals were acquired via a CMS50E fingertip pulse oximeter. During recording, subjects sat approximately 1 meter from the camera and participated in a game designed to induce heart rate variability. The environment was lit by both natural light and a stable artificial light source, resulting in minimal illumination changes—thus considered a static lighting condition.}

\par{COHFACE~\cite{36} comprises video recordings of 40 participants, each recorded under two different lighting conditions: one with good lighting (artificial light turned on) and the other under dim lighting (artificial lights turned off, relying only on ambient natural light). Videos were captured using a Logitech HD Webcam C525 with a resolution of $640\times480$ and a frame rate of 20 fps. Although there is a noticeable difference in lighting intensity between the two setups, the illumination remains constant throughout each video, and the dataset can still be categorized as having static lighting conditions.}

\subsection{rPPG under Complex Lighting Scenarios}

\par{
The rPPG signal captured by cameras from facial videos is inherently weak and becomes increasingly susceptible to noise under low-light conditions, severely compromising measurement accuracy. Study~\cite{14} investigated the extraction of rPPG signals under various low-illumination levels and demonstrated that such environments not only amplify noise but also degrade the performance of chrominance-based algorithms due to the failure of their underlying illumination assumptions.  Moreover, dynamic lighting variations represent another critical factor affecting rPPG performance. Previous studies~\cite{10, 15} introduced handheld device designs that allowed subjects to actively adjust the angle of facial illumination in an effort to simulate more natural lighting variations. However, since these experiments were conducted in controlled environments, the range of lighting changes remained limited. In contrast, study~\cite{5} focused on rPPG measurement in driving scenarios and reported that external occlusions such as trees and buildings can cause rapid changes in facial illumination. This problem becomes especially severe at night, where alternating artificial light sources, such as vehicle headlights and street lamps, result in highly uneven facial lighting. These conditions further increase the likelihood of algorithmic failure. Nevertheless, due to safety concerns, only a limited number of samples were collected, restricting the adaptability and generalizability of deep learning-based rPPG algorithms under dynamic lighting conditions. Therefore, constructing a large-scale rPPG dataset that encompasses diverse and realistic dynamic lighting conditions is both urgent and essential for promoting the practical deployment of rPPG technology in real-world scenarios.
}

\section{Method}
\label{S3Method}

\begin{figure*}[th]
	\centering 
	\includegraphics[width=0.8\textwidth]{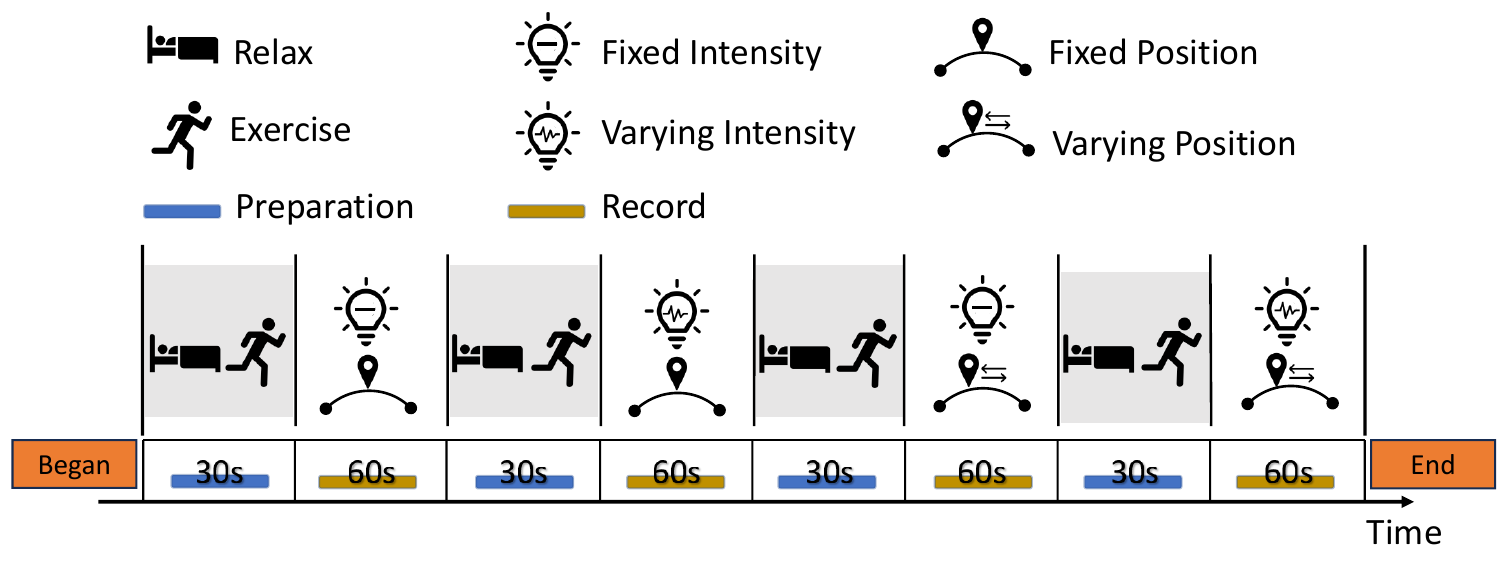}
	\caption{Overview of the DLCN dataset collection process.} 
	\label{2collection} 
\end{figure*}

\begin{figure}[t]
	\centering 
	\includegraphics[width=0.45\textwidth]{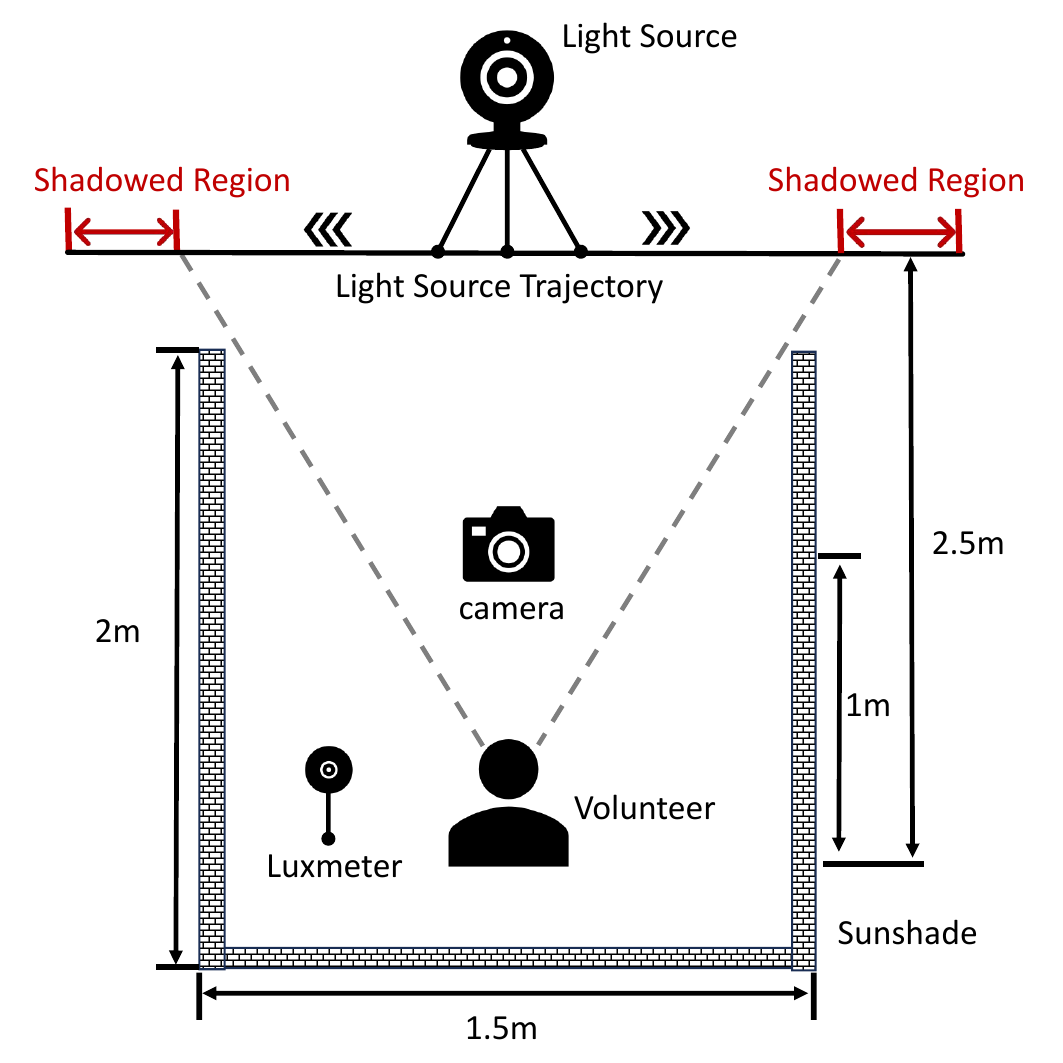}
	\caption{Illustration of the data acquisition setup for the DLCN dataset.} 
	\label{3envoriment} 
\end{figure}

\subsection{DLCN Datasets Collection}
\label{S3.1}
\par{
A total of 98 volunteers (55 males and 43 females) aged between 18 and 30 years were recruited for this study. We collected 784 one-minute video recordings along with synchronized physiological signals under four different lighting conditions. The study protocol was approved by the Medical Ethics Committee of the Seventh People’s Hospital of Chongqing (Approval Number: AF-IRB-RP-015-01). All data collection was conducted in a laboratory setting, and all volunteers provided informed consent after reading and understanding the study procedures.
}

\par{
The overall data collection procedure is illustrated in Fig.~\ref{2collection}. To broaden the heart rate distribution within the dataset, we adopted the strategy described in~\cite{37}, defining two preparation states: a resting state and an active state. The resting state aimed to obtain lower heart rate samples, while the active state (involving exercises such as squats, fast in-place stepping, and jumping jacks) was designed to elicit higher heart rate samples. Each volunteer completed video recordings under the four lighting conditions after each preparation state. The four lighting setups were: (1) fixed intensity and fixed position (FI\&FP), (2) varying intensity and fixed position (VI\&FP), (3) fixed intensity and varying position (FI\&VP), and (4) varying intensity and varying position (VI\&VP).
}

\par{The data collection environment is illustrated in Fig.~\ref{3envoriment}. We constructed a closed blackout booth measuring 2 meters in height, 2 meters in length, and 1.5 meters in width. Except for an opening at the front, all other surfaces were covered with blackout curtains to eliminate external light interference. Volunteers were seated approximately 1 meter from the camera inside the booth, and facial illuminance was measured using a lux meter. A dimmable light source (Panasonic HHLT0663 desk lamp) was positioned about 2.5 meters away from the volunteer on a horizontally movable stand. By adjusting the lamp’s brightness and moving it left or right, we achieved dynamic facial illuminance variations ranging from 10 lux to 100 lux. When the light source was moved to either side, partial facial occlusion of light occurred, simulating real-world light source interference. Video data were captured using a Logitech C922 HD Pro webcam at a resolution of $640\times480$ pixels and a frame rate of 30 fps. Synchronous physiological signals—including PPG, heart rate (HR), and blood oxygen saturation—were recorded using a CONTEC CMS50E fingertip pulse oximeter. We utilized the open-source software PhysRecorder~\cite{38} for synchronized recording of video and physiological signals to ensure high-quality data alignment. Facial illuminance measurements were conducted with a DELIXI DLY-1801C lux meter.}

\begin{figure*}[th]
	\centering 
	\includegraphics[width=\textwidth]{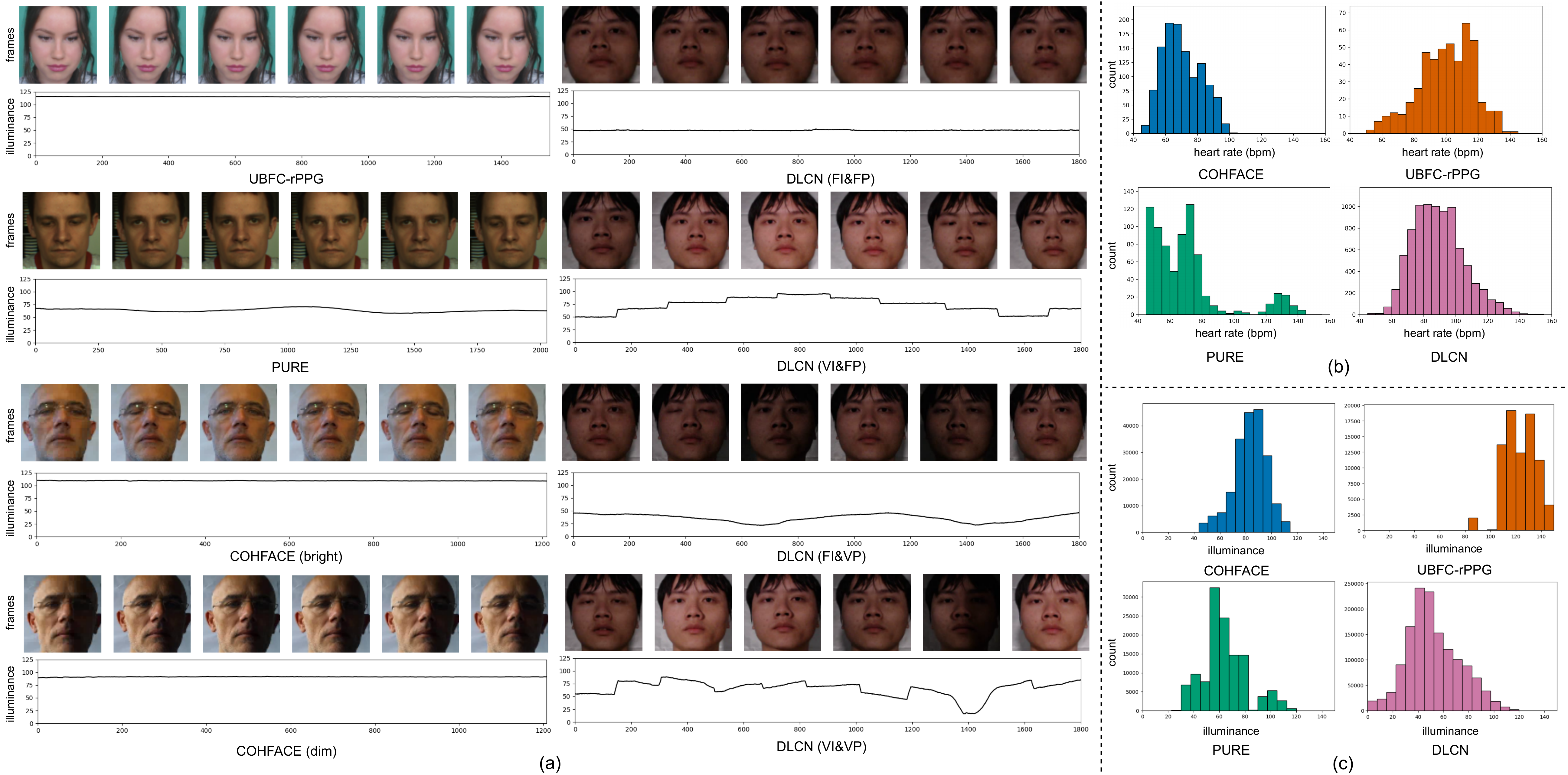}
	\caption{Comparison between the proposed DLCN dataset and three public datasets: UBFC-rPPG~\cite{35}, PURE~\cite{2}, and COHFACE~\cite{36}. (a) Schematic of illumination conditions in each dataset. The left column shows public datasets, while the right shows DLCN. "FI" denotes Fixed Intensity, "FP" denotes Fixed Position, "VI" denotes Varying Intensity, and "VP" denotes Varying Position. (b) Comparison of heart rate distribution across datasets. (c) Comparison of illumination intensity distribution across datasets.} 
	\label{4sample} 
\end{figure*}

\subsection{The challenges of the DLCN}
\label{S3.2}
\subsubsection{Dynamic Illumination Intensity}

\par{Unlike slight illumination fluctuations caused by motion, the DLCN dataset achieves more substantial and diverse illumination intensity variations through controlled adjustments of both light source brightness and position. As shown in Fig.~\ref{4sample}(a), significant differences exist between DLCN and other datasets regarding illumination conditions. In UBFC-rPPG, PURE, and COHFACE datasets, the illumination intensity within individual video samples remains essentially constant. In contrast, DLCN samples exhibit pronounced and diverse changes in illumination. Specifically, the FI\&FP scenario serves as the control group, where facial illumination remains relatively stable but with an overall low brightness level. In the VI\&FP scenario, the light source brightness varies instantaneously over time, resulting in abrupt changes in overall facial brightness. In the FI\&VP scenario, the light source position moves gradually, resulting in continuous but progressive changes in facial illumination intensity. In the VI\&VP scenario, simultaneous variations in both brightness and position further increase the complexity of dynamic facial illumination. Notably, when the light source moves into regions that partially occlude the face, distinct shadow areas appear on the facial region. This design closely simulates real-world, complex lighting conditions, such as nighttime driving or passing through gaps between street lamps, significantly increasing the challenge for models to maintain robustness and generalization in dynamic environments.}

\subsubsection{Broader Heart Rate Distribution}
\par{Fig.~\ref{4sample}(b) illustrates the heart rate distribution across samples from four datasets. The PURE and COHFACE datasets generally exhibit lower heart rate ranges, with COHFACE samples mainly concentrated between 45 and 100 bpm, while PURE’s distribution is narrower, predominantly between 45 and 80 bpm, with only a few samples reaching 120 to 140 bpm. In contrast, the UBFC-rPPG dataset and the proposed DLCN dataset cover a wider range of heart rates. Particularly, the DLCN dataset includes recordings from a post-exercise preparatory state, resulting in a substantial number of high heart rate samples and effectively extending the upper range of the heart rate distribution. Comparatively, the proportion of high heart rate samples in PURE, COHFACE, and UBFC-rPPG is relatively low. This imbalance in heart rate distribution may negatively impact the model’s generalization capability across different heart rate intervals, limiting its adaptability in scenarios with highly dynamic heart rate variations.}

\subsubsection{Lower Illumination Intensity}
\par{We approximate the illumination intensity during data acquisition by calculating the average RGB values of each video frame to represent image brightness. Fig.~\ref{4sample}(c) presents the distribution of illumination intensity across the four datasets based on this metric. It can be observed that the illumination distributions of COHFACE, PURE, and UBFC-rPPG are relatively concentrated, primarily covering medium to high-lighting conditions. Specifically, COHFACE’s illumination intensity is mainly concentrated within the 70–100 lux range with a compact distribution; PURE shows a slightly more dispersed distribution, with most samples falling between 50–80 lux; UBFC-rPPG notably favors higher illumination, with samples predominantly distributed in the 110–140 lux range, and very few samples below 100 lux, indicating limited coverage of low-light environments. In contrast, the DLCN dataset exhibits a more diverse and extensive illumination distribution, especially including a substantial number of samples in the low-light region ($< 60$ lux). Videos captured under such low illumination conditions often suffer from insufficient brightness, low signal-to-noise ratio, and blurred skin features, which pose significant challenges for rPPG signal extraction. The design of DLCN thus enhances its adaptability to complex lighting environments and contributes to improved model robustness in real-world low-light scenarios.}

\begin{figure}[th]
	\centering 
	\includegraphics[width=0.48\textwidth]{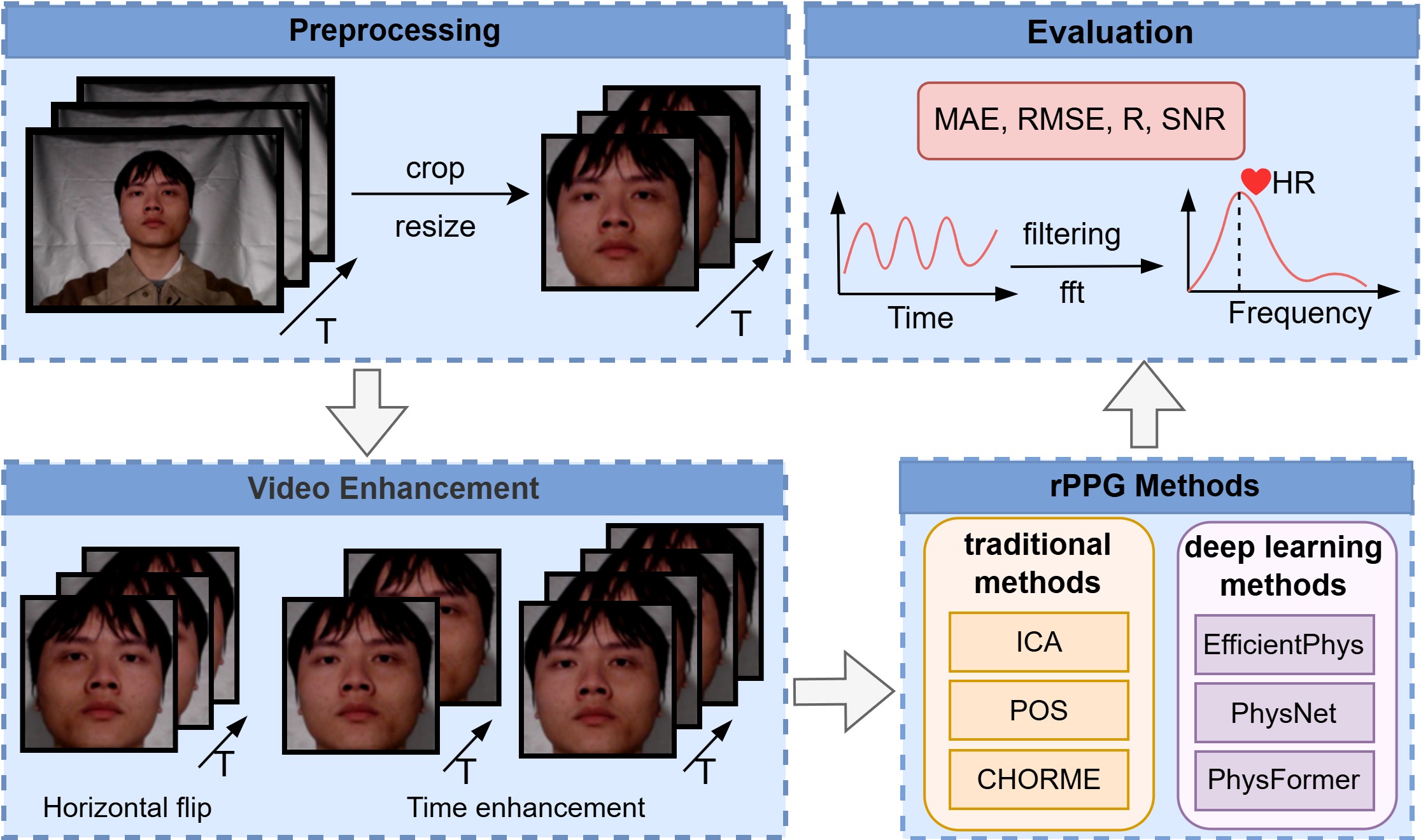}
	\caption{The architecture of the Happy-rPPG Toolkit.} 
	\label{5toolkit} 
\end{figure}

\subsection{The Happy-rPPG Toolkit}
\label{S3.4}
\par{The widely used rPPG toolbox, toolbox-rPPG~\cite{39}, integrates a large number of algorithms, resulting in a relatively complex overall structure and a high degree of encapsulation. This complexity poses challenges for beginners during usage and debugging. To enhance development flexibility and usability, we reconstructed a streamlined and easy-to-debug open-source toolbox named Happy-rPPG Toolkit while retaining partial implementations from toolbox-rPPG. This toolkit incorporates several mainstream deep learning and traditional methods from toolbox-rPPG and additionally provides commonly used modules such as data preprocessing, data augmentation, evaluation metric computation, and result visualization, facilitating users to implement the complete rPPG pipeline. The overall workflow is illustrated in Fig.~\ref{5toolkit}. With the Happy-rPPG Toolkit, researchers can conveniently compare the performance of various algorithms on the proposed DLCN dataset as well as other public datasets, thereby providing strong support for algorithm development and improvement.}

\section{Experiments}
\label{S4Experiments}
\subsection{Implementation Details}
\par{We conducted a unified evaluation of multiple state-of-the-art rPPG methods on the proposed DLCN dataset using the PyTorch-based Happy-rPPG Toolkit. To ensure input consistency, facial landmarks were first extracted from all video frames using the MTCNN algorithm~\cite{40}. The facial region was then cropped with a $1.3\times$ expansion around the landmarks. The cropped face images were then resized to $128\times128$ pixels uniformly. All samples were segmented into non-overlapping clips of 160 frames for both training and testing. Following~\cite{39}, a bandpass filter with a frequency range of 0.75–2.6 Hz was applied in the post-processing stage to the predicted waveform to retain the primary heart rate frequency components. For intra-dataset evaluation, experiments were conducted using five-fold cross-validation. In cross-dataset testing, the models saved from the five-fold training on the source datasets were directly applied to test on other datasets, and the results were averaged. Models were trained for 20 epochs with a batch size of 4 and an initial learning rate of $5\times10^{-5}$. To ensure fairness, all deep learning models were trained using the Negative Pearson correlation loss function~\cite{12}. For traditional rPPG methods, algorithms were directly applied to the entire test set for performance evaluation. All experiments were performed on a Windows operating system platform equipped with an Intel(R) Core(TM) i7-14700KF CPU and an NVIDIA RTX 4090 GPU.}

\subsection{Baseline Methods for Evaluation}
\par{To comprehensively evaluate the performance and applicability of the DLCN dataset across different rPPG algorithms, we selected a diverse set of representative methods covering both traditional and deep learning approaches. For traditional methods, three classic rPPG extraction algorithms were chosen: ICA~\cite{9}, CHROM~\cite{7}, and POS~\cite{8}, which are based on blind source separation and color space transformations, respectively. Additionally, we included several state-of-the-art deep learning models, including EfficientPhys~\cite{11} based on 2D CNNs, PhysNet~\cite{12}, which employs 3D CNNs, and PhysFormer~\cite{13} utilizing a vision transformer architecture. To systematically assess the performance of these algorithms from multiple perspectives, four evaluation metrics were adopted: Mean Absolute Error (MAE), Root Mean Square Error (RMSE), Signal-to-Noise Ratio (SNR), and Pearson Correlation Coefficient ($\rho$), reflecting signal error, noise suppression, and signal correlation, respectively. All algorithms were implemented using the widely adopted toolbox-rPPG~\cite{39} and uniformly integrated into the developed Happy-rPPG Toolkit.}

\begin{table*}[t]
\centering
\caption{Intra-dataset evaluation results. The upper section reports the performance of traditional methods, while the lower section presents the results of deep learning-based approaches.}
\label{table2}
\begin{tabular}{@{}crcccccccccccc@{}}
\toprule
\textbf{} & \multicolumn{1}{c}{\textbf{Method}} & \multicolumn{4}{c}{\textbf{POS~\cite{8}}} & \multicolumn{4}{c}{\textbf{ICA~\cite{9}}} & \multicolumn{4}{c}{\textbf{CHROME~\cite{7}}} \\
\multicolumn{1}{l}{} & \multicolumn{1}{l|}{\textbf{}} & \textbf{MAE ↓} & \textbf{RMSE ↓} & \textbf{$\rho$ ↑} & \multicolumn{1}{c|}{\textbf{SNR ↑}} & \textbf{MAE ↓} & \textbf{RMSE ↓} & \textbf{$\rho$ ↑} & \multicolumn{1}{c|}{\textbf{SNR ↑}} & \textbf{MAE ↓} & \textbf{RMSE ↓} & \textbf{$\rho$ ↑} & \textbf{SNR ↑} \\ \midrule
\multirow{8}{*}{\rotatebox[origin=c]{90}{\textbf{Traditional}}} & \multicolumn{1}{r|}{\textbf{State}} & \multicolumn{1}{l}{} & \multicolumn{1}{l}{} & \multicolumn{1}{l}{} & \multicolumn{1}{l}{} & \multicolumn{1}{l}{} & \multicolumn{1}{l}{} & \multicolumn{1}{l}{} & \multicolumn{1}{l}{} & \multicolumn{1}{l}{} & \multicolumn{1}{l}{} & \multicolumn{1}{l}{} & \multicolumn{1}{l}{} \\
 & \multicolumn{1}{r|}{rest} & 11.016 & 16.173 & 0.300 & \multicolumn{1}{c|}{-8.682} & 18.274 & 24.267 & 0.121 & \multicolumn{1}{c|}{-12.724} & 12.408 & 17.698 & 0.215 & -10.012 \\
 & \multicolumn{1}{r|}{exercise} & 18.089 & 26.094 & 0.141 & \multicolumn{1}{c|}{-12.316} & 26.956 & 35.178 & -0.018 & \multicolumn{1}{c|}{-17.083} & 19.323 & 27.013 & 0.122 & -13.433 \\ \cmidrule(l){2-14} 
 & \multicolumn{1}{r|}{\textbf{Scenario}} &  &  &  &  &  &  &  &  &  &  &  &  \\
 & \multicolumn{1}{r|}{FI\&FP} & 8.042 & 14.030 & 0.550 & \multicolumn{1}{c|}{-4.612} & 9.261 & 17.476 & 0.416 & \multicolumn{1}{c|}{-3.952} & 11.101 & 17.530 & 0.342 & -7.636 \\
 & \multicolumn{1}{r|}{VI\&FP} & 13.608 & 20.581 & 0.246 & \multicolumn{1}{c|}{-9.881} & 24.201 & 30.731 & 0.018 & \multicolumn{1}{c|}{-16.372} & 14.642 & 21.495 & 0.214 & -10.985 \\
 & \multicolumn{1}{r|}{FI\&VP} & 17.275 & 24.035 & 0.107 & \multicolumn{1}{c|}{-12.845} & 27.005 & 33.861 & -0.020 & \multicolumn{1}{c|}{-18.265} & 18.430 & 24.998 & 0.102 & -13.678 \\
 & \multicolumn{1}{r|}{VI\&VP} & 19.192 & 26.113 & 0.039 & \multicolumn{1}{c|}{-14.597} & 29.849 & 35.324 & 0.025 & \multicolumn{1}{c|}{-20.930} & 19.207 & 26.213 & 0.050 & -14.541 \\ \bottomrule

 \noalign{\vskip -4pt}
 
\multicolumn{1}{l}{} & \multicolumn{1}{l}{} & \multicolumn{1}{l}{} & \multicolumn{1}{l}{} & \multicolumn{1}{l}{} & \multicolumn{1}{l}{} & \multicolumn{1}{l}{} & \multicolumn{1}{l}{} & \multicolumn{1}{l}{} & \multicolumn{1}{l}{} & \multicolumn{1}{l}{} & \multicolumn{1}{l}{} & \multicolumn{1}{l}{} & \multicolumn{1}{l}{} \\ 

\toprule
\multicolumn{1}{l}{} & \multicolumn{1}{c|}
{\textbf{Method}} & \multicolumn{4}{c}{\textbf{EfficientPhys~\cite{11}}} & \multicolumn{4}{c}{\textbf{PhysNet~\cite{12}}} & \multicolumn{4}{c}{\textbf{PhysFormer~\cite{13}}} \\
\multicolumn{1}{l}{} & \multicolumn{1}{l|}{\textbf{}} & \textbf{MAE ↓} & \textbf{RMSE ↓} & \textbf{$\rho$ ↑} & \multicolumn{1}{c|}{\textbf{SNR ↑}} & \textbf{MAE ↓} & \textbf{RMSE ↓} & \textbf{$\rho$ ↑} & \multicolumn{1}{c|}{\textbf{SNR ↑}} & \textbf{MAE ↓} & \textbf{RMSE ↓} & \textbf{$\rho$ ↑} & \textbf{SNR ↑} \\ \midrule
\multirow{8}{*}{\rotatebox[origin=c]{90}{\textbf{Deep Learning-based}}} & \multicolumn{1}{r|}{\textbf{State}} & \multicolumn{1}{l}{} & \multicolumn{1}{l}{} & \multicolumn{1}{l}{} & \multicolumn{1}{l}{} & \multicolumn{1}{l}{} & \multicolumn{1}{l}{} & \multicolumn{1}{l}{} & \multicolumn{1}{l}{} & \multicolumn{1}{l}{} & \multicolumn{1}{l}{} & \multicolumn{1}{l}{} & \multicolumn{1}{l}{} \\
 & \multicolumn{1}{r|}{rest} & 5.647 & 11.860 & 0.624 & \multicolumn{1}{c|}{-0.206} & 1.982 & 5.158 & 0.898 & \multicolumn{1}{c|}{6.468} & \textbf{1.699} & \textbf{4.296} & \textbf{0.930} & \textbf{8.866} \\
 & \multicolumn{1}{r|}{exercise} & 9.761 & 20.291 & 0.491 & \multicolumn{1}{c|}{-2.355} & 3.245 & 9.566 & 0.851 & \multicolumn{1}{c|}{4.219} & \textbf{2.482} & \textbf{6.826} & \textbf{0.923} & \textbf{6.581} \\ \cmidrule(l){2-14} 
 & \multicolumn{1}{r|}{\textbf{Scenario}} &  &  &  &  &  &  &  &  & \textbf{} & \textbf{} & \textbf{} & \textbf{} \\
 & \multicolumn{1}{r|}{FI\&FP} & 4.081 & 10.402 & 0.756 & \multicolumn{1}{c|}{1.813} & 1.427 & 4.351 & 0.948 & \multicolumn{1}{c|}{7.345} & \textbf{1.282} & \textbf{3.613} & \textbf{0.965} & \textbf{8.240} \\
 & \multicolumn{1}{r|}{VI\&FP} & 5.824 & 14.052 & 0.601 & \multicolumn{1}{c|}{0.240} & 2.308 & 7.397 & 0.882 & \multicolumn{1}{c|}{6.085} & \textbf{1.714} & \textbf{4.947} & \textbf{0.945} & \textbf{8.402} \\
 & \multicolumn{1}{r|}{FI\&VP} & 11.882 & 21.316 & 0.383 & \multicolumn{1}{c|}{-5.266} & 3.567 & 9.280 & 0.817 & \multicolumn{1}{c|}{3.768} & \textbf{3.293} & \textbf{8.688} & \textbf{0.839} & \textbf{5.985} \\
 & \multicolumn{1}{r|}{VI\&VP} & 13.577 & 23.063 & 0.315 & \multicolumn{1}{c|}{-6.610} & 5.749 & 13.090 & 0.671 & \multicolumn{1}{c|}{0.472} & \textbf{5.343} & \textbf{11.284} & \textbf{0.752} & \textbf{2.088} \\ \bottomrule
\end{tabular}

\vspace{3pt} 
\footnotesize MAE: Mean Absolute Error (bpm), RMSE: Root Mean Square Error (bpm), $\rho$: Pearson Correlation Coefficient, SNR: Signal-to-Noise Ratio (dB).
\end{table*}

\subsection{Intra-dataset Evaluation}

\par{Based on the algorithm implementations within our Happy-rPPG Toolkit, we conducted comprehensive evaluations on the DLCN dataset. To systematically assess the adaptability of different algorithms under various conditions, we performed comparative analyses along two dimensions: (1) resting versus motion states in the data, aimed at investigating the impact of heart rate variations on algorithm performance, and (2) four lighting conditions, to evaluate the stability and robustness of each method under different illumination change scenarios. The experimental results are summarized in Table~\ref{table2}.}

\par{For traditional methods, dynamic lighting variations and low illumination conditions had a significant adverse effect on performance, resulting in generally high errors. Across both resting and motion states, all three traditional algorithms exhibited notably worse performance during motion. Regarding the four lighting scenarios, in the relatively simple FI\&FP condition, the traditional methods demonstrated certain stability, with MAEs of 8.042, 9.261, and 11.101 for POS, ICA, and CHROM, respectively. However, their performance deteriorated markedly in the more challenging VI\&FP and FI\&VP scenarios with substantial lighting changes. Particularly, in the VI\&VP condition, MAEs increased sharply to 19.192 (POS), 29.849 (ICA), and 19.207 (CHROM), indicating that traditional methods almost completely fail under dynamic lighting environments.}

\par{In contrast, deep learning methods consistently outperformed traditional ones across all test scenarios. Among them, PhysFormer showed superior performance compared to EfficientPhys and PhysNet, demonstrating enhanced stability and robustness under complex lighting conditions. It is also observed that the three deep learning models generally perform better in resting states than in motion states. Under the stable illumination condition FI\&FP, the MAEs for EfficientPhys, PhysNet, and PhysFormer were 4.081, 1.427, and 1.282, respectively. Performance declined in the VI\&FP and FI\&VP scenarios with varying light intensities, yet even in the most challenging VI\&VP condition, PhysFormer still achieved an MAE of 5.343, outperforming all other compared methods and showing strong adaptability.}

\begin{table*}[]
\centering
\caption{Cross-dataset evaluation results. Models are trained on UBFC-rPPG, PURE, and COHFACE datasets and tested on the proposed DLCN dataset.}
\label{table3}
\begin{tabular}{@{}crcccccccccccc@{}}
\toprule
\textbf{} & \multicolumn{1}{c}{\textbf{Train}} & \multicolumn{4}{c}{\textbf{UBFC-rPPG~\cite{35}}} & \multicolumn{4}{c}{\textbf{PURE~\cite{2}}} & \multicolumn{4}{c}{\textbf{COHFACE~\cite{36}}} \\
\multicolumn{1}{l}{} & \multicolumn{1}{c|}{\textbf{}} & \textbf{MAE ↓} & \textbf{RMSE ↓} & \textbf{$\rho$ ↑} & \multicolumn{1}{c|}{\textbf{SNR ↑}} & \textbf{MAE ↓} & \textbf{RMSE ↓} & \textbf{$\rho$ ↑} & \multicolumn{1}{c|}{\textbf{SNR ↑}} & \textbf{MAE ↓} & \textbf{RMSE ↓} & \textbf{$\rho$ ↑} & \textbf{SNR ↑} \\ \midrule
\multirow{8}{*}{\rotatebox[origin=c]{90}{\textbf{EfficientPhys~\cite{11}}}} & \multicolumn{1}{r|}{\textbf{State}} &  &  &  &  &  &  &  &  &  &  &  &  \\

 & \multicolumn{1}{r|}{rest} & 11.075 & 16.943 & 0.330 & \multicolumn{1}{c|}{-8.644} & 12.353 & 18.432 & 0.309 & \multicolumn{1}{c|}{-9.121} & 11.995 & 18.371 & 0.318 & -8.665 \\
 & \multicolumn{1}{r|}{exercise} & 16.757 & 25.698 & 0.237 & \multicolumn{1}{c|}{-11.134} & 18.361 & 27.555 & 0.187 & \multicolumn{1}{c|}{-11.870} & 18.241 & 27.738 & 0.204 & -11.403 \\ \cmidrule(l){2-14} 
 & \multicolumn{1}{r|}{\textbf{Scenario}} &  &  &  &  &  &  &  &  &  &  &  &  \\
 & \multicolumn{1}{r|}{FI\&FP} & 5.681 & 12.239 & 0.679 & \multicolumn{1}{c|}{-1.719} & 5.704 & 12.184 & 0.681 & \multicolumn{1}{c|}{-1.664} & 5.514 & 12.271 & 0.685 & -0.999 \\
 & \multicolumn{1}{r|}{VI\&FP} & 14.683 & 21.048 & 0.209 & \multicolumn{1}{c|}{-12.993} & 16.855 & 23.881 & 0.166 & \multicolumn{1}{c|}{-13.707} & 16.497 & 23.494 & 0.173 & -13.690 \\
 & \multicolumn{1}{r|}{FI\&VP} & 15.070 & 23.932 & 0.275 & \multicolumn{1}{c|}{-8.935} & 16.348 & 24.696 & 0.240 & \multicolumn{1}{c|}{-10.179} & 15.772 & 25.022 & 0.275 & -8.720 \\
 & \multicolumn{1}{r|}{VI\&VP} & 20.140 & 26.916 & 0.126 & \multicolumn{1}{c|}{-15.843} & 22.419 & 29.389 & 0.106 & \multicolumn{1}{c|}{-16.361} & 22.586 & 29.653 & 0.110 & -16.657 \\ \midrule
\multirow{8}{*}{\rotatebox[origin=c]{90}{\textbf{PhysNet~\cite{12}}}} & \multicolumn{1}{r|}{\textbf{State}} &  &  &  &  & \multicolumn{1}{l}{} & \multicolumn{1}{l}{} & \multicolumn{1}{l}{} & \multicolumn{1}{l}{} & \multicolumn{1}{l}{} & \multicolumn{1}{l}{} & \multicolumn{1}{l}{} & \multicolumn{1}{l}{} \\
 & \multicolumn{1}{r|}{rest} & 9.716 & 15.442 & 0.375 & \multicolumn{1}{c|}{-6.544} & 8.909 & 14.595 & 0.428 & \multicolumn{1}{c|}{-5.211} & 21.064 & 30.368 & 0.039 & -10.035 \\
 & \multicolumn{1}{r|}{exercise} & 13.346 & 21.884 & 0.332 & \multicolumn{1}{c|}{-8.237} & 13.594 & 22.434 & 0.337 & \multicolumn{1}{c|}{-7.458} & 21.397 & 30.676 & 0.097 & -10.833 \\ \cmidrule(l){2-14} 
 & \multicolumn{1}{r|}{\textbf{Scenario}} &  &  &  &  & \multicolumn{1}{l}{} & \multicolumn{1}{l}{} & \multicolumn{1}{l}{} & \multicolumn{1}{l}{} & \multicolumn{1}{l}{} & \multicolumn{1}{l}{} & \multicolumn{1}{l}{} & \multicolumn{1}{l}{} \\
 & \multicolumn{1}{r|}{FI\&FP} & 5.630 & 11.721 & 0.690 & \multicolumn{1}{c|}{-0.277} & 4.689 & 10.595 & 0.745 & \multicolumn{1}{c|}{0.698} & 23.426 & 34.528 & 0.106 & -7.215 \\
 & \multicolumn{1}{r|}{VI\&FP} & 12.596 & 19.810 & 0.282 & \multicolumn{1}{c|}{-8.433} & 10.588 & 17.788 & 0.399 & \multicolumn{1}{c|}{-6.399} & 18.197 & 25.791 & 0.080 & -11.283 \\
 & \multicolumn{1}{r|}{FI\&VP} & 10.936 & 17.929 & 0.429 & \multicolumn{1}{c|}{-7.202} & 13.083 & 20.729 & 0.320 & \multicolumn{1}{c|}{-7.822} & 24.308 & 34.139 & 0.076 & -9.978 \\
 & \multicolumn{1}{r|}{VI\&VP} & 16.619 & 23.967 & 0.189 & \multicolumn{1}{c|}{-12.172} & 16.570 & 23.880 & 0.194 & \multicolumn{1}{c|}{-11.758} & 19.000 & 25.811 & 0.108 & -13.236 \\ \midrule
\multirow{8}{*}{\rotatebox[origin=c]{90}{\textbf{PhysFormer~\cite{13}}}} & \multicolumn{1}{r|}{\textbf{State}} & \multicolumn{1}{l}{} & \multicolumn{1}{l}{} & \multicolumn{1}{l}{} & \multicolumn{1}{l}{} & \multicolumn{1}{l}{} & \multicolumn{1}{l}{} & \multicolumn{1}{l}{} & \multicolumn{1}{l}{} & \multicolumn{1}{l}{} & \multicolumn{1}{l}{} & \multicolumn{1}{l}{} & \multicolumn{1}{l}{} \\
 & \multicolumn{1}{r|}{rest} & \textbf{7.637} & \textbf{13.060} & \textbf{0.505} & \multicolumn{1}{c|}{\textbf{-3.392}} & 9.600 & 15.470 & 0.387 & \multicolumn{1}{c|}{-5.627} & 11.223 & 20.732 & 0.271 & -2.628 \\
 & \multicolumn{1}{r|}{exercise} & \textbf{12.028} & \textbf{20.745} & \textbf{0.408} & \multicolumn{1}{c|}{\textbf{-5.882}} & 15.466 & 24.437 & 0.261 & \multicolumn{1}{c|}{-8.523} & 12.331 & 22.097 & 0.366 & -4.207 \\ \cmidrule(l){2-14}  
 & \multicolumn{1}{r|}{\textbf{Scenario}} & \multicolumn{1}{l}{} & \multicolumn{1}{l}{} & \multicolumn{1}{l}{} & \multicolumn{1}{l}{} & \multicolumn{1}{l}{} & \multicolumn{1}{l}{} & \multicolumn{1}{l}{} & \multicolumn{1}{l}{} & \multicolumn{1}{l}{} & \multicolumn{1}{l}{} & \multicolumn{1}{l}{} & \multicolumn{1}{l}{} \\
 & \multicolumn{1}{r|}{FI\&FP} & \textbf{3.944} & \textbf{9.192} & \textbf{0.802} & \multicolumn{1}{c|}{\textbf{1.990}} & 5.475 & 11.763 & 0.695 & \multicolumn{1}{c|}{-0.198} & 10.476 & 23.321 & 0.397 & 2.070 \\
 & \multicolumn{1}{r|}{VI\&FP} & \textbf{9.872} & \textbf{16.679} & \textbf{0.440} & \multicolumn{1}{c|}{\textbf{-5.021}} & 11.289 & 18.805 & 0.364 & \multicolumn{1}{c|}{-6.416} & 10.408 & 18.652 & 0.358 & -3.334 \\
 & \multicolumn{1}{r|}{FI\&VP} & \textbf{10.837} & \textbf{18.760} & \textbf{0.415} & \multicolumn{1}{c|}{\textbf{-5.258}} & 15.255 & 23.127 & 0.196 & \multicolumn{1}{c|}{-9.142} & 12.170 & 22.212 & 0.367 & -3.107 \\
 & \multicolumn{1}{r|}{VI\&VP} & \textbf{14.611} & \textbf{21.984} & \textbf{0.276} & \multicolumn{1}{c|}{\textbf{-10.202}} & 18.028 & 25.326 & 0.158 & \multicolumn{1}{c|}{-12.483} & 14.038 & 21.167 & 0.291 & -9.256 \\ \bottomrule
\end{tabular}

\vspace{3pt} 
\footnotesize MAE: Mean Absolute Error (bpm), RMSE: Root Mean Square Error (bpm), $\rho$: Pearson Correlation Coefficient, SNR: Signal-to-Noise Ratio (dB).
\end{table*}

\subsection{Cross-dataset Evaluation}
\par{We evaluated the generalization capability of deep learning methods across different DLCN scenarios by training models separately on the UBFC-rPPG, PURE, and COHFACE datasets and testing on the DLCN dataset. Consistent with the intra-dataset evaluation, the DLCN data were partitioned based on preparation states and lighting conditions. The experimental results are summarized in Table~\ref{table3}. Overall, across both resting and motion preparation states, all methods exhibited better cross-dataset performance during resting states compared to motion states. Regarding the four lighting conditions, the simplest scenario, FI\&FP, yielded the best performance. In scenarios involving changes in light intensity or source position (VI\&FP and FI\&VP), model performance is significantly degraded. The lowest performance was observed in the most complex scenario, VI\&VP, where both light intensity and source position varied simultaneously. Among the three methods, EfficientPhys demonstrated relatively consistent results when trained on the three source datasets. PhysNet performed better when trained on PURE compared to training on the other two datasets. Notably, models trained on COHFACE exhibited the poorest generalization to DLCN, with performance nearly failing entirely. In contrast, PhysFormer trained on UBFC-rPPG achieved the best results on DLCN, attaining an MAE of 3.944 in the FI\&FP scenario, outperforming all other compared methods.}

\begin{figure*}[th]
	\centering 
	\includegraphics[width=\textwidth]{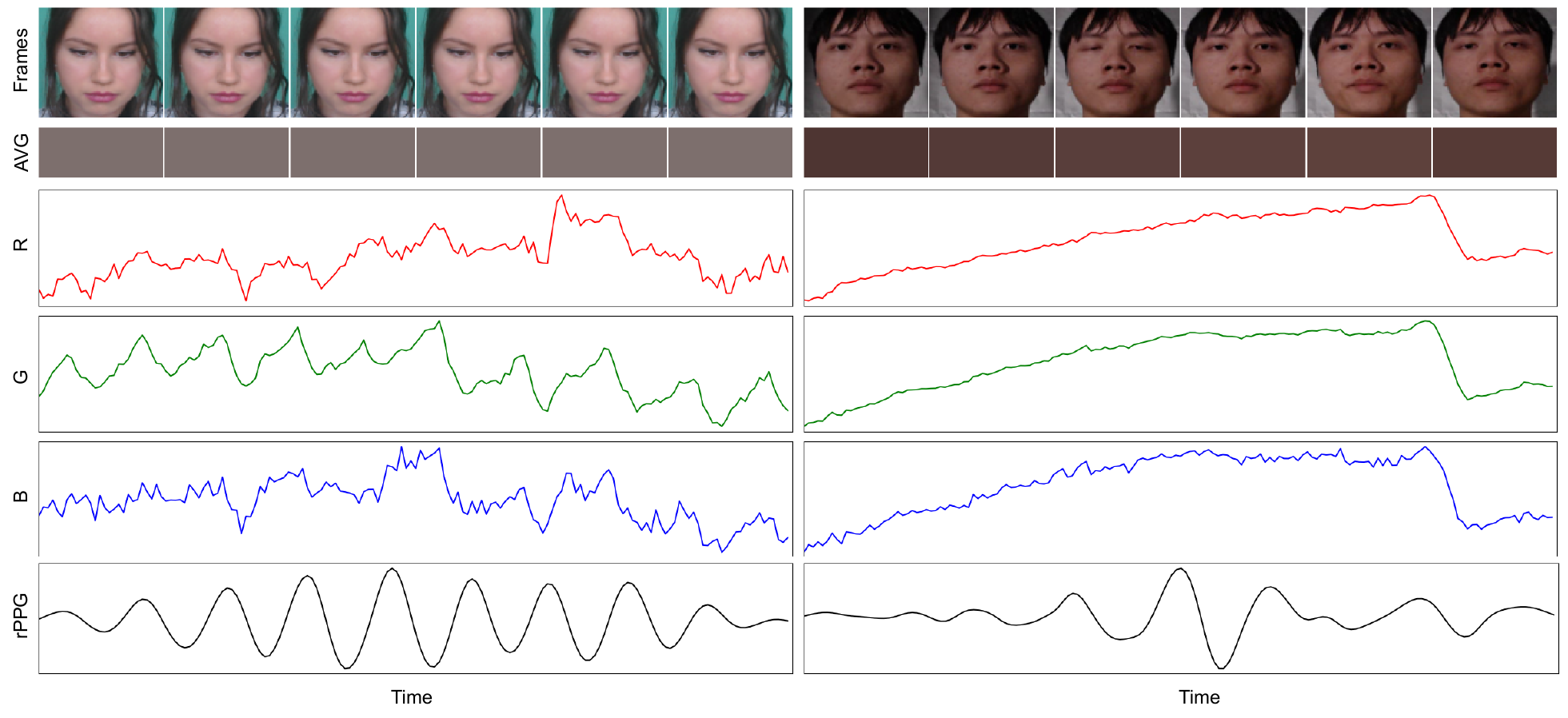}
	\caption{Visualization of temporal variation in mean facial RGB values under simple and dynamic lighting conditions. The left plot shows a sample from UBFC-rPPG with stable lighting, while the right plot shows a DLCN sample under dynamic illumination. UBFC-rPPG samples exhibit clear periodicity in facial RGB signals, facilitating rPPG extraction. In contrast, DLCN samples exhibit significant disruption to periodicity due to lighting variations, which increases the difficulty of signal extraction.} 
	\label{6challenge1} 
\end{figure*}

\begin{table*}[]
\centering
\caption{Cross-scenario evaluation within the DLCN dataset. Models are trained on the FI\&FP scenario and tested on VI\&FP, FI\&VP, and VI\&VP scenarios, respectively.}
\label{table4}
\begin{tabular}{@{}ccccccccccccc@{}}
\toprule
\textbf{Method} & \multicolumn{4}{c}{\textbf{FI\&FP $\rightarrow$ VI\&FP}} & \multicolumn{4}{c}{\textbf{FI\&FP $\rightarrow$  FI\&VP}} & \multicolumn{4}{c}{\textbf{FI\&FP  $\rightarrow$  VI\&VP}} \\
\multicolumn{1}{c|}{\textbf{}} & \textbf{MAE ↓} & \textbf{RMSE ↓} & \textbf{$\rho$ ↑} & \multicolumn{1}{c|}{\textbf{SNR ↑}} & \textbf{MAE ↓} & \textbf{RMSE ↓} & \textbf{$\rho$ ↑} & \multicolumn{1}{c|}{\textbf{SNR ↑}} & \textbf{MAE ↓} & \textbf{RMSE ↓} & \textbf{$\rho$ ↑} & \textbf{SNR ↑} \\ \midrule
\multicolumn{1}{c|}{EfficientPhys~\cite{11}} & 14.236 & 21.858 & 0.262 & \multicolumn{1}{c|}{-10.510} & 14.026 & 23.996 & 0.314 & \multicolumn{1}{c|}{-6.736} & 22.086 & 29.786 & 0.106 & -15.644 \\
\multicolumn{1}{c|}{PhysNet~\cite{12}} & 8.423 & 15.347 & 0.506 & \multicolumn{1}{c|}{-3.457} & 7.121 & 15.133 & 0.584 & \multicolumn{1}{c|}{-0.768} & 15.081 & 22.741 & 0.240 & -9.913 \\
\multicolumn{1}{c|}{PhysFormer~\cite{13}} & 9.752 & 16.466 & 0.456 & \multicolumn{1}{c|}{-4.217} & 8.350 & 16.521 & 0.548 & \multicolumn{1}{c|}{-1.474} & 15.604 & 22.600 & 0.274 & -11.694 \\ \bottomrule
\end{tabular}

\vspace{3pt} 
\footnotesize MAE: Mean Absolute Error (bpm), RMSE: Root Mean Square Error (bpm), $\rho$: Pearson Correlation Coefficient, SNR: Signal-to-Noise Ratio (dB).
\end{table*}

\section{Discussion}
\label{S5Discussion}
\subsection{Challenges under Dynamic Lighting Conditions}
\par{
From the intra-dataset experimental results presented in Table~\ref{table2}, it is evident that traditional algorithms suffer a significant performance drop on the DLCN dataset, indicating their limited adaptability to complex lighting and state variations. Traditional methods typically estimate the pulse signal based on changes in the average RGB values of the facial region; however, under dynamic lighting scenarios, the variation patterns of RGB values become more complex, leading to the failure of these methods (as illustrated in Fig.~\ref{6challenge1}). In contrast, data-driven deep learning approaches demonstrate stronger robustness under dynamic lighting conditions. Nevertheless, their performance still noticeably degrades in more complex dynamic lighting scenarios, such as VI\&FP, FI\&VP, and VI\&VP. Cross-dataset test results in Table~\ref{table3} further show that models trained on public datasets UBFC-rPPG, PURE, and COHFACE generally perform poorly on the DLCN dataset’s dynamic lighting scenarios, except for the simplest FI\&FP scenario. This indicates that the lighting conditions covered by existing datasets are insufficient to support effective generalization of models in real-world complex lighting environments.
}

\par{
To further analyze the challenges explicitly posed by dynamic lighting itself, excluding the confounding effects of heart rate distribution and illumination levels, we designed a controlled variable experiment using the DLCN dataset. Models were trained under the FI\&FP scenario and tested on the VI\&FP, FI\&VP, and VI\&VP scenarios. As shown in Table~\ref{table4}, in the relatively simpler VI\&FP and FI\&VP scenarios, both PhysNet and PhysFormer achieve MAEs around 8; however, in the most complex VI\&VP scenario, their MAEs increase to approximately 15. By comparison, EfficientPhys exhibits MAEs exceeding 14 across all three scenarios. These results collectively indicate that current methods still struggle to maintain stable generalization performance under dynamic lighting changes, highlighting the necessity of improving algorithmic adaptability in such conditions as a key direction for future research.
}

\begin{table*}[]
\centering
\caption{Impact of temporal augmentation (TA) on model performance.}
\label{table5}
\begin{tabular}{cccccccccc}
\toprule
\textbf{Method} & \textbf{w/ TA} & \multicolumn{4}{c}{\textbf{PURE $\rightarrow$ DLCN (FI\&FP)}} & \multicolumn{4}{c}{\textbf{DLCN (FI\&FP) $\rightarrow$ PURE}} \\
\textbf{} & \multicolumn{1}{c|}{} & \textbf{MAE ↓} & \textbf{RMSE ↓} & \textbf{$\rho$ ↑} & \multicolumn{1}{c|}{\textbf{SNR ↑}} & \textbf{MAE ↓} & \textbf{RMSE ↓} & \textbf{$\rho$ ↑} & \textbf{SNR ↑} \\ \midrule
\multirow{2}{*}{EfficientPhys~\cite{11}} & \multicolumn{1}{c|}{$\times$} & 5.842 & 12.483 & 0.666 & \multicolumn{1}{c|}{-1.677} & 8.27 & 17.858 & 0.634 & -0.726 \\
 & \multicolumn{1}{c|}{\checkmark} & 5.704 & 12.184 & 0.681 & \multicolumn{1}{c|}{-1.664} & 7.964 & 17.53 & 0.644 & -0.459 \\ \midrule
\multirow{2}{*}{PhysNet~\cite{12}} & \multicolumn{1}{c|}{$\times$} & 10.990 & 19.084 & 0.268 & \multicolumn{1}{c|}{-4.574} & 5.931 & 15.591 & 0.714 & 4.366 \\
 & \multicolumn{1}{c|}{\checkmark} & \textbf{4.689} & \textbf{10.595} & \textbf{0.745} & \multicolumn{1}{c|}{\textbf{0.698}} & \textbf{4.954} & \textbf{13.919} & \textbf{0.782} & \textbf{5.357} \\ \midrule
\multirow{2}{*}{PhysFormer~\cite{13}} & \multicolumn{1}{c|}{$\times$} & 8.061 & 16.136 & 0.457 & \multicolumn{1}{c|}{-1.582} & 6.864 & 15.572 & 0.744 & 4.116 \\
 & \multicolumn{1}{c|}{\checkmark} & 5.475 & 11.763 & 0.695 & \multicolumn{1}{c|}{-0.198} & 5.791 & 14.291 & 0.78 & 4.835 \\ \bottomrule
\end{tabular}

\vspace{3pt}
\footnotesize MAE: Mean Absolute Error (bpm), RMSE: Root Mean Square Error (bpm), $\rho$: Pearson Correlation Coefficient, SNR: Signal-to-Noise Ratio (dB).
\end{table*}

\begin{table*}[]
\centering
\caption{Impact of temporal normalization on model generalization performance.}
\label{table6}
\begin{tabular}{cccccccccc}
\toprule
\textbf{Method} & \textbf{w/ TN} & \multicolumn{4}{c}{\textbf{UBFC-rPPG $\rightarrow$ DLCN (FI\&FP)}} & \multicolumn{4}{c}{\textbf{DLCN (FI\&FP) $\rightarrow$ UBFC-rPPG}} \\
\textbf{} & \multicolumn{1}{c|}{} & \textbf{MAE ↓} & \textbf{RMSE ↓} & \textbf{$\rho$ ↑} & \multicolumn{1}{c|}{\textbf{SNR ↑}} & \textbf{MAE ↓} & \textbf{RMSE ↓} & \textbf{$\rho$ ↑} & \textbf{SNR ↑} \\ \midrule

\multirow{2}{*}{EfficientPhys~\cite{11}} 
 & \multicolumn{1}{c|}{$\times$} & 5.833 & 12.658 & 0.660 & \multicolumn{1}{c|}{-1.350} & 8.213 & 18.240 & 0.602 & -2.361 \\
 & \multicolumn{1}{c|}{$\checkmark$} & 5.681 & 12.239 & 0.679 & \multicolumn{1}{c|}{-1.719} & 5.254 & 14.189 & 0.721 & 0.179 \\ \midrule

\multirow{2}{*}{PhysNet~\cite{12}}
 & \multicolumn{1}{c|}{$\times$} & 5.913 & 11.773 & 0.677 & \multicolumn{1}{c|}{-1.708} & 7.696 & 17.741 & 0.573 & -0.379 \\
 & \multicolumn{1}{c|}{$\checkmark$} & 5.630 & 11.721 & 0.690 & \multicolumn{1}{c|}{-0.277} & 4.045 & 11.473 & 0.799 & 2.709 \\ \midrule

\multirow{2}{*}{PhysFormer~\cite{13}}
 & \multicolumn{1}{c|}{$\times$} & 6.666 & 12.836 & 0.620 & \multicolumn{1}{c|}{-2.041} & 5.043 & 13.022 & 0.758 & 3.304 \\
 & \multicolumn{1}{c|}{$\checkmark$} & \textbf{3.944} & \textbf{9.192} & \textbf{0.802} & \multicolumn{1}{c|}{\textbf{1.990}} & \textbf{2.221} & \textbf{6.013} & \textbf{0.940} & \textbf{5.952} \\ 
\bottomrule
\end{tabular}

\vspace{3pt}
\footnotesize MAE: Mean Absolute Error (bpm), RMSE: Root Mean Square Error (bpm), $\rho$: Pearson Correlation Coefficient, SNR: Signal-to-Noise Ratio (dB).
\end{table*}

\subsection{Challenges under High Heart Rate}
\par{
From the intra-dataset test results in Table~\ref{table2}, it is evident that all traditional and deep learning methods perform worse in the exercise state compared to the resting state, indicating that high heart rate scenarios during physical activity pose significant challenges for extracting the rPPG signal. Additionally, cross-dataset test results in Table~\ref{table3} reveal that models trained on the UBFC-rPPG dataset generally outperform those trained on PURE and COHFACE when tested across datasets. By examining the heart rate distribution shown in Fig.~\ref{4sample}(b), it can be observed that the heart rate distributions of UBFC-rPPG and DLCN datasets are relatively similar. In contrast, PURE and COHFACE contain noticeably fewer samples with high heart rates. This imbalance in sample distribution leads to reduced model generalization capability, a conclusion consistent with findings reported in~\cite{30, 37, 41}.
}

\begin{figure}[t]
	\centering 
	\includegraphics[width=0.48\textwidth]{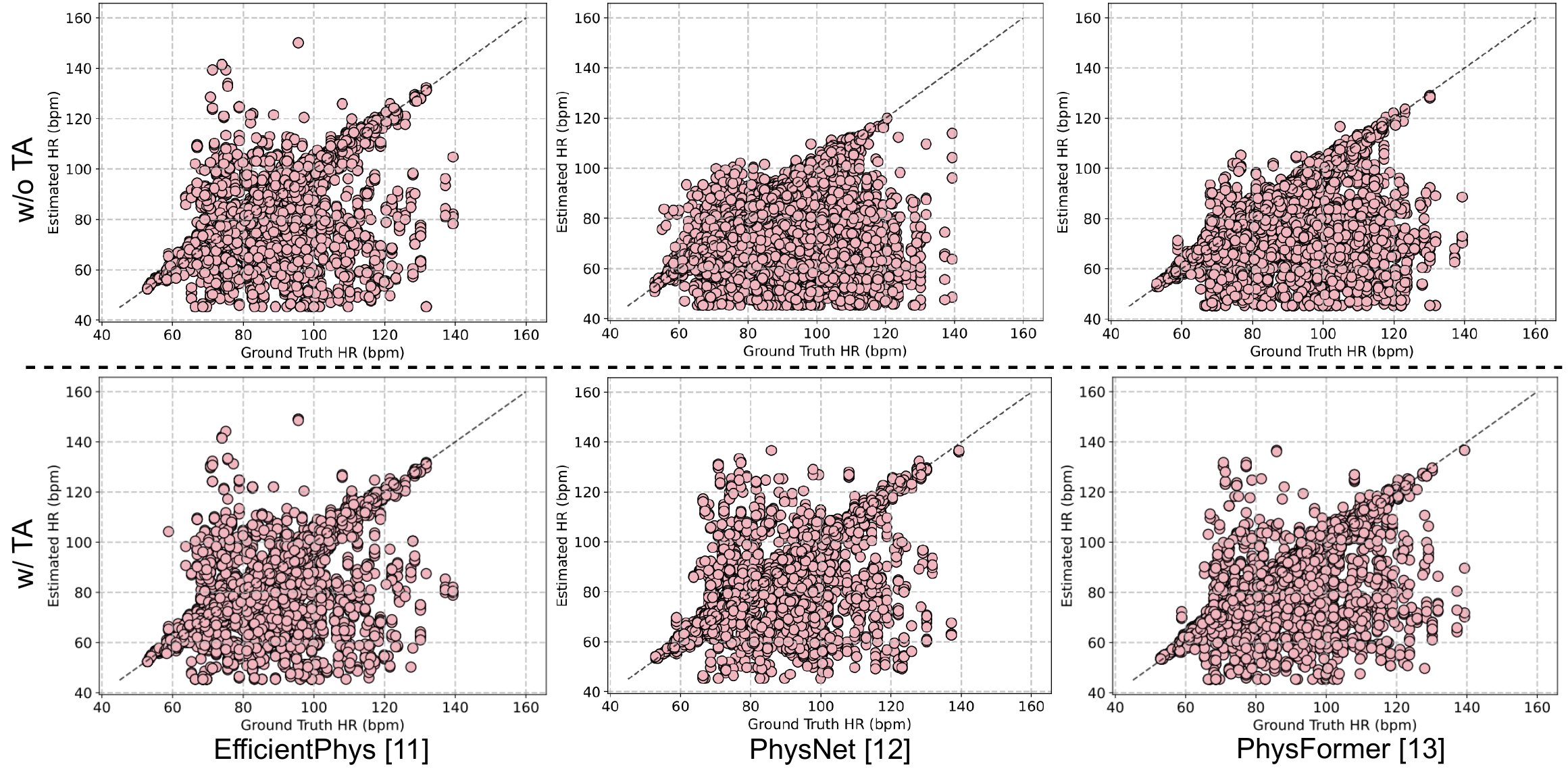}
	\caption{Scatter plot of ablation experiments on temporal augmentation (TA). The model is trained on the PURE dataset and tested on the DLCN (FI\&FP) scenario. Results show the effect of TA in enhancing high heart rate prediction.} 
	\label{7challenge2} 
\end{figure}

\par{
To mitigate the generalization issues caused by differences in heart rate distributions, we adopt the temporal augmentation strategy introduced in~\cite{13} and~\cite{30}. Experiments were conducted on DLCN (FI\&VP) and PURE datasets, which exhibit significant disparities in heart rate distributions. Results in Table~\ref{table5} demonstrate that temporal augmentation substantially improves the generalization performance of all three methods when tested on datasets with distribution mismatches. Taking as an example the test results on DLCN (FI\&FP) after training on PURE (visualized in the scatter plot of Fig.~\ref{7challenge2}), temporal augmentation notably increases the number of high heart rate samples correctly predicted by the model. These findings suggest that temporal augmentation enriches the heart rate distribution in training samples, thereby effectively enhancing the model’s generalization capability.
}
\begin{figure*}[t]
	\centering 
	\includegraphics[width=\textwidth]{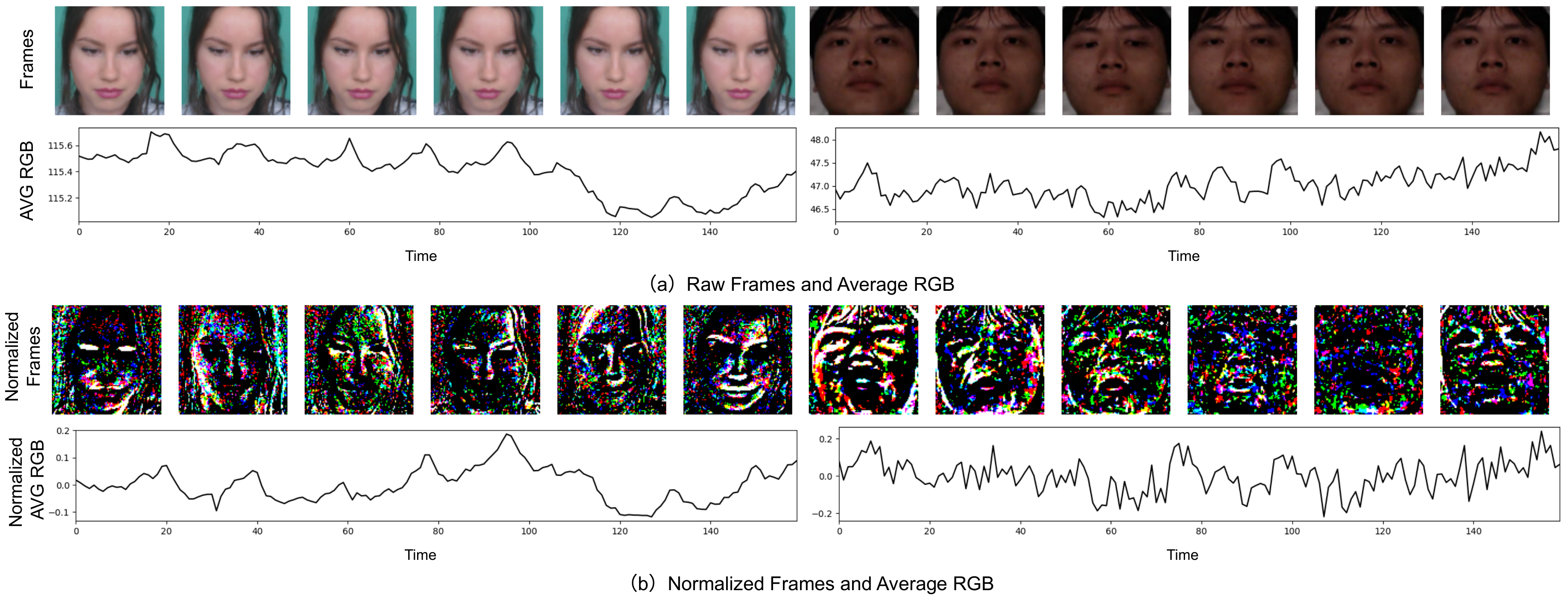}
	\caption{Visualization of normalized samples under high and low illumination. After normalization, brightness differences between samples are significantly reduced while preserving original temporal brightness trends.} 
	\label{8challenge3} 
\end{figure*}

\subsection{Challenges under Low Illumination Conditions}
\par{
Variations in illumination intensity result in inconsistent input data distributions for models. To alleviate the negative impact of such distribution discrepancies on model generalization, we adopt the normalization method proposed in~\cite{41}, which normalizes input data to ensure that the input distributions under different lighting conditions are consistent during validation. Considering the significant differences in illumination intensity between the UBFC-rPPG and DLCN datasets, while their heart rate distributions remain relatively similar, we select the DLCN FI\&FP scenario and the UBFC-rPPG dataset for cross-dataset evaluation to analyze the influence of illumination conditions. The experimental results are presented in Table~\ref{table6}. The results indicate that after applying temporal normalization, the model’s generalization performance improves for both the high illumination UBFC-rPPG dataset and the low illumination DLCN (FI\&FP) scenario.
}
\par{
Additionally, we visualize the normalized samples (see Fig.~\ref{8challenge3}). The visualization shows that temporal normalization significantly reduces the differences in brightness among samples under varying illumination intensities while preserving the overall brightness variation trends. This normalization operation effectively maps input samples from different lighting conditions into a unified range, thereby enhancing the model’s generalization capability.
}

\subsection{Application under dynamic light conditions at night}
\par{
It is worth emphasizing that dynamic illumination conditions at night are very common in real-world rPPG applications, especially in safety-critical tasks such as driver monitoring~\cite{5}. With the rapid development of intelligent driving technologies, safety monitoring during nighttime driving remains a major challenge before these technologies become fully mature. If rPPG technology can be reliably applied under such dynamic nighttime environments, it holds great promise for significantly enhancing driving safety and reducing the risk of potential accidents~\cite{42}.
}

\subsection{Limitations and Future Work}
\par{
Compared to other high-quality public datasets (e.g., VIPL~\cite{15}, MMPD~\cite{10}), the current version of the DLCN dataset lacks systematic coverage of variables such as motion states and individuals with diverse skin tones, which may limit its generalization capabilities across broader populations and real-world scenarios. Moreover, real-world lighting environments are far more complex than the four designed conditions in our dataset, often involving multiple light sources, colored illumination, and frequently fluctuating light intensities. In future work, we plan to continuously expand the scale and diversity of the DLCN dataset by incorporating more complex lighting and motion scenarios. Additionally, we aim to develop more robust nighttime rPPG signal extraction methods based on this dataset to further promote the practical deployment of rPPG technology in real-world applications such as nighttime driving.
}

\section{Conclusion}
\label{S6Conclusion}
\par{
The proposed DLCN dataset provides a more challenging benchmark for evaluating rPPG algorithms under nighttime dynamic lighting conditions, offering significant research and practical value. Comprehensive experiments were conducted on the DLCN dataset using the developed Happy-rPPG Toolkit. The results indicate that traditional methods almost completely fail under nighttime dynamic lighting, whereas data-driven deep-learning approaches still demonstrate a certain degree of stability in such complex scenarios. Enhancing model generalization across diverse nighttime dynamic lighting conditions remains a key issue for advancing the real-world application of rPPG technology. The main challenges lie in dynamically varying lighting environments, discrepancies in heart rate distribution across samples, and low illumination intensity—all of which are critical problems that future research must address.
}


 \bibliographystyle{IEEEtran}
\bibliography{refs}

\vfill

\end{document}